\documentclass[10pt, a4paper]{article}

\usepackage{lrec-coling2024} 
\usepackage{tabularx}
\usepackage{booktabs}

\usepackage{appendix}

\title{Branching Narratives: Character Decision Points Detection}

\name{Alexey Tikhonov} 

\address{Inworld.AI \\
         Berlin, Germany \\
         altsoph@gmail.com \\}

\abstract{
This paper presents the Character Decision Points Detection (CHADPOD) task, a task of identification of points within narratives where characters make decisions that may significantly influence the story's direction. 
We propose a novel dataset based on Choose Your Own Adventure (a registered trademark of Chooseco LLC) games graphs to be used as a benchmark for such a task. We provide a comparative analysis of different models' performance on this task, including a couple of LLMs and several MLMs as baselines, achieving up to 89\% accuracy. This underscores the complexity of narrative analysis, showing the challenges associated with understanding character-driven story dynamics.
Additionally, we show how such a model can be applied to the existing text to produce linear segments divided by potential branching points, demonstrating the practical application of our findings in narrative analysis. 
 \\ \newline \Keywords{NLP, narrative analysis, CYOA, agency} }

\begin{document}

\maketitleabstract

\section{Introduction}

Modern Large Language Models (LLMs) are state-of-the-art in a lot of Natural Language Processing (NLP) tasks. However, areas related to the analysis and generation of texts with complex and rich semantic structures remain underexplored. This includes the tasks of analyzing and generating long, engaging, and rich narratives \cite{van-stegeren-theune-2019-narrative}. While modern models can sometimes produce innovative plot twists, they generally create less imaginative scenarios and rhetoric compared to human-authored texts \cite{begus2023experimental}.

The traditional machine learning approach to this problem starts from the data collection  with necessary annotations. In the narrative analysis field, there are a number of datasets available, such as WikiPlots\footnote{https://github.com/markriedl/WikiPlots} with 112,936 story plots extracted from English Wikipedia, the MPST dataset with 14K movie plot synopses \cite{kar2018mpst}, and the DYPLODOC dataset, which includes synopses of 13K TV shows, 21K seasons, and over 300K episodes \cite{Malysheva_2021}. However, these plain-text synopses offer limited assistance when the goal is to analyze high-level narrative structures.

Philosophers and linguists make a lot of attempts to conceptualize and formalize concepts of plot, narrative arcs, character development, conflict, and so on \cite{shklovsky1925theory}. One of the fundamental principles in drama and narrative construction is the concept of character agency, which posits that a character's decisions and actions drive the plot forward. \citeauthor{Philosophy}, in his work \textit{Poetics}, highlights sudden plot twists, or \textit{peripeteia}, especially those tied to \textit{anagnorisis}—the moment when a character comes to a significant realization or discovery that affects subsequent choices. \citeauthor{borges1964labyrinths}, in his short story \textit{The Garden of Forking Paths}, explores the idea of multiple possible worlds through the metaphor of a labyrinth, representing an infinite number of potential narratives and outcomes based on characters' actions: "\textit{
your ancestor
<...> believed in an infinite series of times, in a growing, dizzying net of divergent, convergent and parallel times. This web of time—the strands of which approach one another, bifurcate, intersect or ignore each other through the centuries—embrace every possibility}." \citeauthor{propp1968morphology} introduces the concept of functions—recurring, typical actions that move the narrative forward: "\textit{a tale often
attributes identical actions to various personages; this makes
possible the study of the tale according to the functions of its
dramatis personae <...> a function <...> cannot be defined apart
from its place in the course of narration}". Gustav Freytag, in \cite{freytag1968freytag}, describes \textit{Freytag's Pyramid}, a typical plot structure identifying five pivotal plot points: Opportunity, Change of Plans, Point of No Return, Major Setback, Climax. \citeauthor{aarseth1997cybertext} in his \textit{Cybertext} book proposes the term \textit{ergodic literature} to define open, dynamic texts, with which the reader must perform specific actions to generate a literary sequence.

One may argue that we still cannot clearly define what we aim to analyze, and this slows progress in the analysis and generation of narrative structures \cite{yamshchikov-tikhonov-2023-wrong}. However, the NLP community continues to seek improvements in narrative processing \cite{fan2019strategies}, by setting subtasks for the formal identification of important plot elements. For instance, in \cite{tikhonov2022actionable} the task is to identify "Chekhov's guns"—narrative objects that significantly impact the plot's development; \cite{papalampidi-etal-2019-movie} introduce a Turning Point Identification task—to directly identify Freytag's points in the text, and \cite{li2023automated} proposes a task to extract action models from narrative texts automatically.

In this paper, we propose using characters' decision-making moments to analyze and formalize narrative structure. We introduce a new NLP task—Character Decision Points Detection (CHADPOD). This task focuses on identifying moments in the narrative where characters make decisions that significantly determine the plot's direction. We believe that highlighting such moments will improve our understanding of traditional text plots and open possibilities for working with nonlinear and interactive narrative structures \cite{juul2005half, murray2006toward}.

This work contributes by:
\begin{enumerate}
\item Proposing a formalization of the Character Decision Points Detection (CHADPOD) task.
\item Introducing a Character Decision Points dataset.
\item Demonstrating the effectiveness of modern models in identifying Character Decision Points (CDPs).
\item Offering an interpretation of CDPs and their relation to the related task of turning points.
\end{enumerate}

\section{CHADPOD task}

In NLP research, analysts and creators frequently utilize Gamebook genre games, also widely known as Choose Your Own Adventure\footnote{It is a registered trademark of Chooseco LLC.} (CYOA) books, named after one of the earliest popular series in this genre. These sources are crucial for studying nonlinear narratives, alongside interactive fiction games.

For instance, the data from such sources has been used to train systems that generate suggestions for people writing short stories \cite{clark-smith-2021-choose}. Another study employs CYOA as a medium for training generative agents to enforce temporal constraints \cite{rothkopf2024enforcing}. In the MACHIAVELLI paper \cite{pan2023rewards}, authors use a collection of CYOA games to create a game environment for training text agents. Some researchers\footnote{https://heterogenoustasks.wordpress.com/2015\\/01/26/standard-patterns-in-choice-based-games/} explore them to analyze a variety of narrative macro-structures.

In this work, we introduce the CHADPOD task, which focuses on identifying narrative points where a character makes a choice that determines the further course of the story. We utilize CYOA game graphs to create a new CHADPOD benchmark, consisting of 1,462 binary classification tasks, with 731 tasks in each class. Each task comprises two text segments—a prefix and a postfix. The positive class includes narrative points where a character makes a choice that significantly influences the story's direction. The negative class consists of randomly segmented texts (we take a continuous text from a single node and split it at some random point between sentences), as well as text points where a character takes some action, but it does not significantly affect the story's progression.

\section{Data}

\begin{figure*}[!h]
\begin{center}
\includegraphics[scale=0.35]{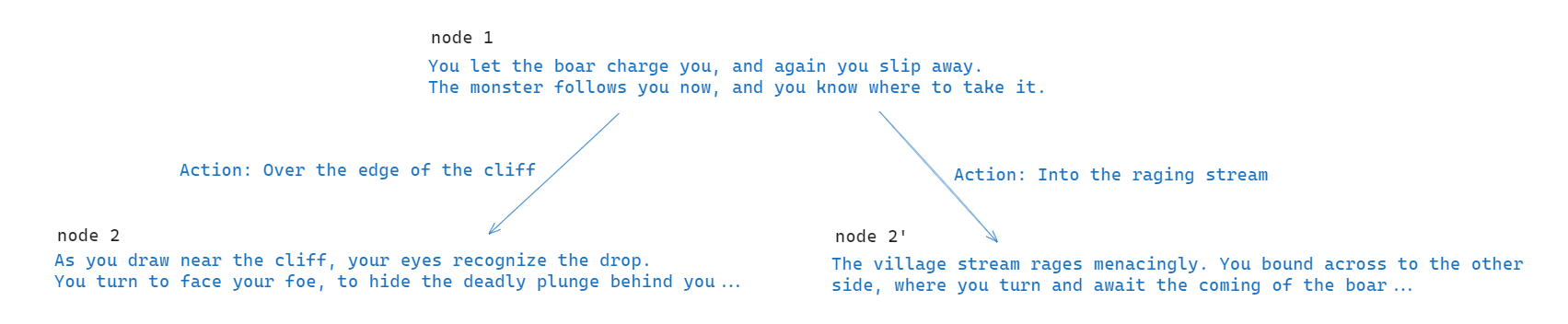} 
\caption{Example of branching in CYOA data, shortened for the simplicity.}
\label{fig.graph}
\end{center}
\end{figure*}

In this section, we describe the process of constructing the CHADPOD dataset.

We use the MACHIAVELLI dataset \cite{pan2023rewards}, which consists of 134 Choose-Your-Own-Adventure games, as our input data.

For each available game, we analyze its graph and extract triplets of the form:
\begin{center}<\textit{node1}; \textit{action}; \textit{node2}>\end{center}
where \textit{node1} is the text before the action, \textit{action} is the choice made by the player, and \textit{node2} is the text following the action (see Figure \ref{fig.graph}).

Next, we filter the triplets—removing exact duplicates, retaining only those with descriptions sufficiently long  to provide enough context -- both in \textit{node1} and \textit{node2} (to do so we used simple heuristics -- at least 4 sentences, at least 50 characters), removing texts that are dialogue segments (dialogues are a very specific type of narrative that should be analyzed separately, see for example \cite{zhou2023dialogue}), and removing texts with unusual characters. As positive examples of branching points, we only select triplets for which the graph from node1 has more than one possible action, thus excluding scenarios like <\textit{node1}, “1 year later…”, \textit{node2}>. The remaining 731 examples make up the positive class.

Then we form the negative class from two components—using the division of texts from the same games (nodes) at random points as easy negatives, and the above-described cases when there is exactly one action emanating from node1 in the graph as hard negatives.

Finally, we divide the data\footnote{The data is available through \href{https://drive.google.com/file/d/1_xfryK7Ku7yO79mBcsPoTaK4pJmsWnMM/view}{Google Drive}. The password is CHADPOD.} into 3 game-wise splits, ensuring that there are no overlaps between the splits in terms of games, thereby eliminating test set leakage. The statistics of the resulting split are presented in Table \ref{tab:dataset_split}.

\begin{table}[ht]
\centering
\caption{Data Splits}
\begin{tabular}{lccc}
\hline
\textbf{Class} & \textbf{Train} & \textbf{Dev} & \textbf{Test} \\
\hline
Positives     & 511   & 110 & 110  \\
Negatives     & 256   & 55  & 55   \\
Hard Negatives& 255   & 55  & 55   \\
\textbf{Total}         & 1022  & 220 & 220  \\
\hline
\end{tabular}
\label{tab:dataset_split}
\end{table}

\section{Task validation}
\paragraph{Experiments}
To validate the usefulness of our dataset, we trained several models for the CHADPOD tasks. We used the DeBERTa model \cite{he2021deberta} as a strong baseline, known for its state-of-the-art performance in many text classification tasks \footnote{\url{https://huggingface.co/altsoph/chadpod}}. Additionally, we included older but widely used models such as BERT \cite{devlin-etal-2019-bert} and ALBERT \cite{lan2020albert} as weaker baselines. We chose accuracy as the metric due to our data being class-balanced. The training was conducted on a GPU RTX 4090 on the Vast.ai platform, with each model trained with a batch size of 4 and a learning rate of \(5.5 \times 10^{-6}\) until accuracy on a validation split began to decline. A full training run for one task required at most 30 minutes. We also added results for GPT-3.5-turbo\footnote{https://openai.com/blog/gpt-3-5-turbo-fine-tuning-and-api-updates} and GPT-4-turbo\footnote{https://openai.com/blog/new-models-and-developer-products-announced-at-devday} tested in a zero-shot manner with hyperparamteres (temperature, probability threshold) obtained by a grid search on the validation set. 

The results are presented in the Table \ref{tab:model_accuracy}.

\begin{table}[ht]
\centering
\caption{\small{Test Accuracy of Models on CHADPOD}}
\begin{tabular}{lcc}
\hline
\textbf{model} & \textbf{test acc} & \textbf{size} \\
\hline
\small{DeBERTa-v3-large} & \small{\textbf{89\%}} & \small{340M} \\
\small{DeBERTa-v3-base} & \small{85\%} & \small{110M} \\
\small{ALBERT-v2-base} & \small{84\%} & \small{11M} \\
\small{BERT-base} & \small{79\%} & \small{110M} \\
\small{GPT-4-turbo, 0-shot} & \small{62\%} & \small{unknown} \\
\small{GPT-3.5-turbo, 0-shot} & \small{55\%} & \small{unknown} \\
\hline
\end{tabular}
\label{tab:model_accuracy}
\end{table}

As seen, the task is solvable but remains quite complex for simpler and smaller models. The presented results are on a test dataset without overlap with the training set in terms of games, minimizing the risk of overfitting.
As for LLMs, it seems that using them in a 0-shot manner is not a silver bullet for this task, though results could likely be improved through fine-tuning or prompt engineering. 
Analysis of the confusion matrix revealed that LLMs underperform on the positive class, leading to a high number of false negatives.


\paragraph{Comparison with Turning Points}

One may notice that the CHADPOD task is significantly similar to the Turning Points Identification task proposed in \cite{papalampidi-etal-2019-movie}. In this section, we conduct a comparative analysis, demonstrating that despite similar formulations, the tasks differ fundamentally.

Recall that in \cite{papalampidi-etal-2019-movie}, the TRIPOD dataset consists of manually annotated short plot synopses (averaging 35 sentences) of 99 screenplays with sentence-level turning points annotations, where turning points are defined as the 5 classic pivot moments formulated in Freytag's Pyramid.

We transformed the TRIPOD dataset to our format, taking contexts around the indicated turning points as positive examples and random divisions of the same synopses where there were no turning points as negative examples. The final dataset used all available non-overlapping contexts with at least 3 sentences before and after the split point, resulting in 255 positive and 209 negative examples.

Applying our DeBERTa-v3-large based model to these examples yielded the metrics provided in Table \ref{tab:tripod_performance}.

\begin{table}[ht]
\centering
\caption{Performance on Adapted TRIPOD Dataset}
\begin{tabular}{lcc}
\hline
\textbf{Metric} & \textbf{Value} \\
\hline
Accuracy & 40\% \\
Balanced Accuracy & 41\% \\
F1-Score & 41\% \\
\hline
\end{tabular}
\label{tab:tripod_performance}
\end{table}

These results indicate that the semantics of the tasks significantly differ (recall that the model's accuracy on an isolated test set was 89\%).

One might suggest that the main difference between these tasks lies in the scale (turning points are just 5 key moments in the plot's macrostructure) and in that Freytag's turning points do not necessarily imply character agency. Contrarily, they can be exclusively formed by external events, leaving characters without a choice.



\section{Text Segmentation Study}

In this part, we demonstrate how the obtained binary classification model can be used for segmenting text into linear segments separated by potential branching points in the narrative.

\begin{figure}[!ht]
\begin{center}
\includegraphics[scale=0.35]{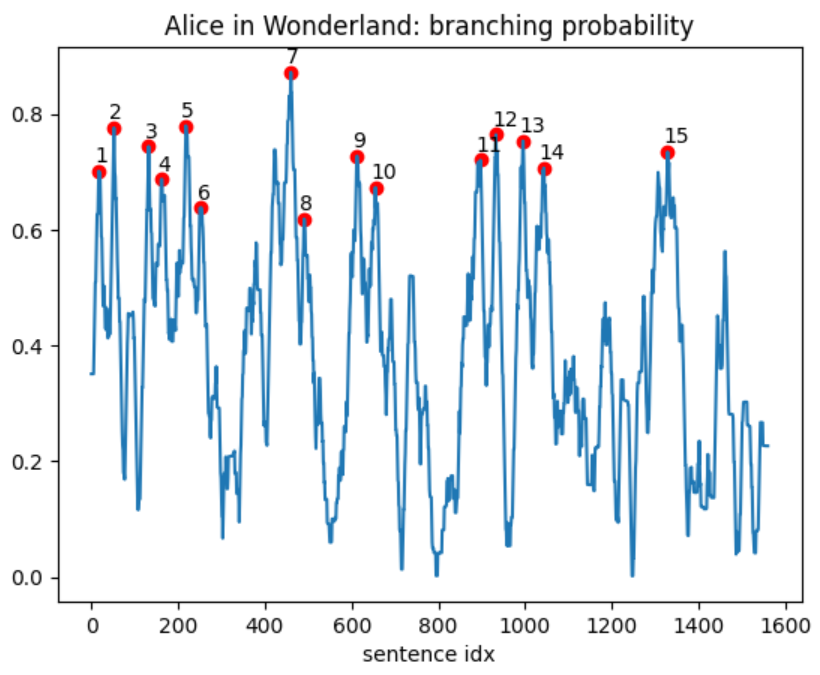} 
\caption{Most probable branching points in the text of Alice in Wonderland.}
\label{fig.alice}
\end{center}
\end{figure}

For our experiments, we utilized the text of \textit{Alice's Adventures in Wonderland} by Lewis Carroll, as it is in the public domain. We employed a sliding window of 10 sentences with a step of 1 sentence and calculated the probability of branching at any given possible point. To reduce noise, we applied a convolution with a linear kernel of width 25, and then on the resulting sequence, we identified local maximums on segments lying above the threshold \(TH1=0.5\), preserving only peaks above the threshold \(TH2=0.6\) to obtain 15 main branching points. Indeed, the parameters of such a heuristic can be adjusted to change the sensitivity of the approach.

In the Figure \ref{fig.alice} one can see 15 most probable points of branching in the given text. To analyse them and gain understanding if these points really correspond to the important decisions of the chraracter, we used GPT-4-turbo model to assess selected points and propose potential alternatives to the character's action. We refer to the Table \ref{tab:alice} for details of these points and alternatives. 

Despite the subjectivity of such analysis, it is worth saying that all identified points, except maybe for 1 and 4, correspond to moments when a character makes a decision or performs an action that influences the subsequent development of events.

\section{Discussion}

This study contributes to the evolving field of Natural Language Processing (NLP) by addressing the nuanced task of detecting Character Decision Points (CDPs) within narrative texts. Through the development and validation of the CHADPOD task, our findings highlight the complexity and potential of leveraging Large Language Models (LLMs) for narrative analysis, particularly in identifying moments of narrative branching that may be important to story development.

The performance of various models on the CHADPOD task, especially the high  results of the DeBERTa model, demonstrates the feasibility of detecting narrative branching points with high accuracy. However, the underperformance of smaller and simpler models, as well as zero-shot tests of GPT-3.5 and GPT-4, illustrates the challenges of the task. We suggest these challenges are not solely due to model capacity but also reflect the sophisticated understanding of narrative structure.


The application of our binary classification model to text segmentation, as demonstrated in the analysis of Alice's Adventures in Wonderland, showcases the practical utility of our approach. This illustrative study can be a bit speculative without ground truth labeling, since ChatGPT is able to generate plausible alternatives for any requested point in text. However, subjectively, most of the detected branching points (demonstrated in Table \ref{tab:alice}) correspond to the turning points of text there the character makes impactful decisions.
(this problem can also be approached as a direct segmentation task, as in, for example, \cite{koshorek2018text}; we leave these experiments for future work.)

Our results suggest several directions for future research. First, expanding the dataset to include a broader range of narratives, 
could enhance the model's understanding of diverse narrative structures. Second, exploring more granular classifications of decision points, such as presented in Syd Field's book \textit{Screenplay} (with 6 key points) or the one based on Vogler's interpretation of Campbell's monomyth (with 12 such points) 
could offer finer insights into narrative dynamics. Third, using the CHADPOD data can help to construct a macro-assessment of characters' agency within a text, i.e., an assessment that enables comparing different texts in terms of how much the development of the text is determined by the characters' choices. \\\\\\\\

\nocite{*}
\pagebreak
\section{Bibliographical References}\label{sec:reference}

\bibliographystyle{lrec-coling2024-natbib}
\bibliography{lrec-coling2024-example}

\begin{thebibliography}{25}
\expandafter\ifx\csname natexlab\endcsname\relax\def\natexlab#1{#1}\fi

\bibitem[{Aarseth(1997)}]{aarseth1997cybertext}
E.J. Aarseth. 1997.
\newblock \emph{Cybertext: perspectives on ergodic literature}.
\newblock Johns Hopkins Univ Pr.

\bibitem[{Aristotle(2014)}]{Philosophy}
Aristotle. 2014.
\newblock \emph{Aristotle collection}.
\newblock Annotated Classics.

\bibitem[{Begus(2023)}]{begus2023experimental}
Nina Begus. 2023.
\newblock \href {http://arxiv.org/abs/2310.12902} {Experimental narratives: A
  comparison of human crowdsourced storytelling and ai storytelling}.

\bibitem[{Borges(1964)}]{borges1964labyrinths}
Jorge~Luis Borges. 1964.
\newblock \emph{Labyrinths: selected stories \& other writings}, volume 186.
\newblock New Directions Publishing.

\bibitem[{Clark and Smith(2021)}]{clark-smith-2021-choose}
Elizabeth Clark and Noah~A. Smith. 2021.
\newblock \href {https://doi.org/10.18653/v1/2021.naacl-main.279} {Choose your
  own adventure: Paired suggestions in collaborative writing for evaluating
  story generation models}.
\newblock In \emph{Proceedings of NAACL 2021}, pages 3566--3575, Online.
  Association for Computational Linguistics.

\bibitem[{Devlin et~al.(2019)Devlin, Chang, Lee, and
  Toutanova}]{devlin-etal-2019-bert}
Jacob Devlin, Ming-Wei Chang, Kenton Lee, and Kristina Toutanova. 2019.
\newblock \href {https://doi.org/10.18653/v1/N19-1423} {{BERT}: Pre-training of
  deep bidirectional transformers for language understanding}.
\newblock In \emph{Proceedings of NAACL 2019}, pages 4171--4186, Minneapolis,
  Minnesota. Association for Computational Linguistics.

\bibitem[{Fan et~al.(2019)Fan, Lewis, and Dauphin}]{fan2019strategies}
Angela Fan, Mike Lewis, and Yann Dauphin. 2019.
\newblock Strategies for structuring story generation.
\newblock \emph{arXiv preprint arXiv:1902.01109}.

\bibitem[{Freytag and MacEwan(1968)}]{freytag1968freytag}
G.~Freytag and E.J. MacEwan. 1968.
\newblock \href {https://books.google.de/books?id=kbRUyAEACAAJ}
  {\emph{Freytag's Technique of the Drama: An Exposition of Dramatic
  Composition and Art}}.
\newblock Johnson Reprint.

\bibitem[{He et~al.(2021)He, Liu, Gao, and Chen}]{he2021deberta}
Pengcheng He, Xiaodong Liu, Jianfeng Gao, and Weizhu Chen. 2021.
\newblock \href {http://arxiv.org/abs/2006.03654} {Deberta: Decoding-enhanced
  bert with disentangled attention}.

\bibitem[{Juul(2005)}]{juul2005half}
Jesper Juul. 2005.
\newblock Half-real: Video games between real rules and fictional worlds.

\bibitem[{Kar et~al.(2018)Kar, Maharjan, López-Monroy, and
  Solorio}]{kar2018mpst}
Sudipta Kar, Suraj Maharjan, A.~Pastor López-Monroy, and Thamar Solorio. 2018.
\newblock \href {http://arxiv.org/abs/1802.07858} {Mpst: A corpus of movie plot
  synopses with tags}.

\bibitem[{Koshorek et~al.(2018)Koshorek, Cohen, Mor, Rotman, and
  Berant}]{koshorek2018text}
Omri Koshorek, Adir Cohen, Noam Mor, Michael Rotman, and Jonathan Berant. 2018.
\newblock \href {http://arxiv.org/abs/1803.09337} {Text segmentation as a
  supervised learning task}.

\bibitem[{Lan et~al.(2020)Lan, Chen, Goodman, Gimpel, Sharma, and
  Soricut}]{lan2020albert}
Zhenzhong Lan, Mingda Chen, Sebastian Goodman, Kevin Gimpel, Piyush Sharma, and
  Radu Soricut. 2020.
\newblock \href {http://arxiv.org/abs/1909.11942} {Albert: A lite bert for
  self-supervised learning of language representations}.

\bibitem[{Li et~al.(2023)Li, Cui, Lin, and Haslum}]{li2023automated}
Ruiqi Li, Leyang Cui, Songtuan Lin, and Patrik Haslum. 2023.
\newblock \href {http://arxiv.org/abs/2307.10247} {Automated action model
  acquisition from narrative texts}.

\bibitem[{Malysheva et~al.(2021)Malysheva, Tikhonov, and
  Yamshchikov}]{Malysheva_2021}
Anastasia Malysheva, Alexey Tikhonov, and Ivan~P. Yamshchikov. 2021.
\newblock \href {https://doi.org/10.3233/faia210283} {\emph{DYPLODOC: Dynamic
  Plots for Document Classification}}. IOS Press.

\bibitem[{Murray(2006)}]{murray2006toward}
Janet~H Murray. 2006.
\newblock Toward a cultural theory of gaming: Digital games and the
  co-evolution of media, mind, and culture.
\newblock \emph{Popular Communication}, 4(3):185--202.

\bibitem[{Pan et~al.(2023)Pan, Chan, Zou, Li, Basart, Woodside, Ng, Zhang,
  Emmons, and Hendrycks}]{pan2023rewards}
Alexander Pan, Jun~Shern Chan, Andy Zou, Nathaniel Li, Steven Basart, Thomas
  Woodside, Jonathan Ng, Hanlin Zhang, Scott Emmons, and Dan Hendrycks. 2023.
\newblock \href {http://arxiv.org/abs/2304.03279} {Do the rewards justify the
  means? measuring trade-offs between rewards and ethical behavior in the
  machiavelli benchmark}.

\bibitem[{Papalampidi et~al.(2019)Papalampidi, Keller, and
  Lapata}]{papalampidi-etal-2019-movie}
Pinelopi Papalampidi, Frank Keller, and Mirella Lapata. 2019.
\newblock \href {https://doi.org/10.18653/v1/D19-1180} {Movie plot analysis via
  turning point identification}.
\newblock In \emph{Proceedings of the 2019 Conference on Empirical Methods in
  Natural Language Processing and the 9th International Joint Conference on
  Natural Language Processing (EMNLP-IJCNLP)}, pages 1707--1717, Hong Kong,
  China. Association for Computational Linguistics.

\bibitem[{Propp(1968)}]{propp1968morphology}
Vladimir Propp. 1968.
\newblock Morphology of the folktale, trans.
\newblock \emph{Louis Wagner, 2d. ed.}

\bibitem[{Rothkopf et~al.(2024)Rothkopf, Zeng, and
  Santolucito}]{rothkopf2024enforcing}
Raven Rothkopf, Hannah~Tongxin Zeng, and Mark Santolucito. 2024.
\newblock \href {http://arxiv.org/abs/2402.16905} {Enforcing temporal
  constraints on generative agent behavior with reactive synthesis}.

\bibitem[{Shklovsky(1925)}]{shklovsky1925theory}
Viktor Shklovsky. 1925.
\newblock Theory of prose (b. sher, trans.).
\newblock \emph{Champaign, IL: Dalkey Archive Press. Original work published}.

\bibitem[{Tikhonov and Yamshchikov(2022)}]{tikhonov2022actionable}
Alexey Tikhonov and Ivan~P. Yamshchikov. 2022.
\newblock \href {http://arxiv.org/abs/2109.13855} {Actionable entities
  recognition benchmark for interactive fiction}.

\bibitem[{van Stegeren and Theune(2019)}]{van-stegeren-theune-2019-narrative}
Judith van Stegeren and Mari{\"e}t Theune. 2019.
\newblock \href {https://doi.org/10.18653/v1/W19-3407} {{N}arrative
  {G}eneration in the {W}ild: {M}ethods from {N}a{N}o{G}en{M}o}.
\newblock In \emph{Proceedings of the Second Workshop on Storytelling}, pages
  65--74, Florence, Italy. Association for Computational Linguistics.

\bibitem[{Yamshchikov and Tikhonov(2023)}]{yamshchikov-tikhonov-2023-wrong}
Ivan Yamshchikov and Alexey Tikhonov. 2023.
\newblock \href {https://doi.org/10.18653/v1/2023.wnu-1.8} {What is wrong with
  language models that can not tell a story?}
\newblock In \emph{Proceedings of the The 5th Workshop on Narrative
  Understanding}, pages 58--64, Toronto, Canada. Association for Computational
  Linguistics.

\bibitem[{Zhou et~al.(2023)Zhou, Peng, and Riedl}]{zhou2023dialogue}
Wei Zhou, Xiangyu Peng, and Mark Riedl. 2023.
\newblock \href {http://arxiv.org/abs/2307.15833} {Dialogue shaping: Empowering
  agents through npc interaction}.

\end{thebibliography}


\newpage

\begin{table*}[b]
\centering
\caption{Branching Points and Alternatives in \textit{Alice's Adventures in Wonderland}}
\fontsize{8}{10}\selectfont
\begin{tabularx}{\textwidth}{clX}
\toprule
\textbf{No.} & \textbf{Main Decision Point} & \textbf{Alternatives} \\
\midrule
1 & Alice falls down the well & \begin{tabular}[c]{@{}l@{}}- Tries to look for something to grab onto to stop\\ - Attempts to fly or float by flapping her arms\end{tabular} \\
\midrule
2 & Alice uses the little golden key to open the small door & \begin{tabular}[c]{@{}l@{}}- Tries breaking one of the doors with a chair\\ - Climbs onto the table\end{tabular} \\
\midrule
3 & Alice cries a pool of tears and falls there & \begin{tabular}[c]{@{}l@{}}- Calls out for help\\ - Eats something to change her size\end{tabular} \\
\midrule
4 & Alice thinks she is Mable and continues to cry & \begin{tabular}[c]{@{}l@{}}- Searches for someone who remember her\\ - Insists that she is not Mabel\end{tabular} \\
\midrule
5 & Alice tells Mice about dogs and scares it & \begin{tabular}[c]{@{}l@{}}- Stops talking about pets\\ - Apologizes to the Mouse\end{tabular} \\
\midrule
6 & Alice decides to join the Caucus-race & \begin{tabular}[c]{@{}l@{}}- Suggests a different activity\\ - Objects to the Caucus-race\end{tabular} \\
\midrule
7 & Alice decides to look on top of the mushroom & \begin{tabular}[c]{@{}l@{}}- Eats a flower\\ - Goes back to the puppy\end{tabular} \\
\midrule
8 & Alice agrees to return to the Caterpillar & \begin{tabular}[c]{@{}l@{}}- Ignores it and walks away\\ - Loses her temper with the Caterpillar\end{tabular} \\
\midrule
9 & Alice knocks on the door & \begin{tabular}[c]{@{}l@{}}- Searches for another entrance\\ - Returns to the wood\end{tabular} \\
\midrule
10 & Alice tries to calm down the cook & \begin{tabular}[c]{@{}l@{}}- Leaves the room\\ - Organizes a cleanup effort\end{tabular} \\
\midrule
11 & Alice decides to leave the tea-party forever & \begin{tabular}[c]{@{}l@{}}- Stays despite rudeness\\ - Invites the Dormouse to leave\end{tabular} \\
\midrule
12 & Alice stays standing on the arrival of King and Queen & \begin{tabular}[c]{@{}l@{}}- Lies down like the gardeners\\ - Starts clapping\end{tabular} \\
\midrule
13 & Alice decides to talk with the Cat about the game & \begin{tabular}[c]{@{}l@{}}- Ignores the Cat and plays alone\\ - Attempts to leave the croquet ground\end{tabular} \\
\midrule
14 & Alice argues with the Duchess & \begin{tabular}[c]{@{}l@{}}- Agrees with the Duchess\\ - Changes the subject\end{tabular} \\
\midrule
15 & Alice follows the Gryphon to the trial & \begin{tabular}[c]{@{}l@{}}- Returns to the Mock Turtle\\ - Stays to listen to the soup song\end{tabular} \\
\bottomrule
\end{tabularx}
\label{tab:alice}
\end{table*}

\end{document}